\documentclass[review]{elsarticle}
\usepackage{amsmath,amssymb}
\usepackage{algorithmic}
\usepackage{graphicx}
\usepackage{multirow}
\usepackage{algorithm}
\usepackage{algorithmic}
\usepackage{arydshln} 
\usepackage{lineno,hyperref}
\journal{Journal of \LaTeX\ Templates}
\begin{document}

\begin{frontmatter}

\title{Multi-source adversarial transfer learning based on similar source domains with local features}
%% or include affiliations in footnotes:
\author[mymainaddress]{Yifu Zhang}
\ead{2000912@stu.neu.edu.cn}
%\ead[url]{www.elsevier.com}
\author[mymainaddress]{Hongru Li\corref{mycorrespondingauthor}}
\cortext[mycorrespondingauthor]{Correspondence to: College of Information Sciences and Engineering, Northeastern University, NO. 3-11, Wenhua Road, Heping District, Shenyang, 110819, China.}
\ead{lihongru@ise.neu.edu.cn}
\author[mymainaddress]{Shimeng Shi}
\ead{2100819@stu.neu.edu.cn}
\author[mymainaddress]{Youqi Li}
\ead{2270751@stu.neu.edu.cn}
\author[mymainaddress2]{Jiansong Zhang}
\ead{786064222@qq.com}
\address[mymainaddress]{College of Information Sciences and Engineering, Northeastern University, Shenyang, 110819, China}
\address[mymainaddress2]{Henan Educational Resource Guarantee Center, Zhengzhou 450004, China}

\begin{abstract}
Transfer learning leverages knowledge from other domains and has been successful in many applications. Transfer learning methods rely on the overall similarity of the source and target domains. However, in some cases, it is impossible to provide an overall similar source domain, and only some source domains with similar local features can be provided. Can transfer learning be achieved? In this regard, we propose a multi-source adversarial transfer learning method based on local feature similarity to the source domain to handle transfer scenarios where the source and target domains have only local similarities. This method extracts transferable local features between a single source domain and the target domain through a sub-network. Specifically, the feature extractor of the sub-network is induced by the domain discriminator to learn transferable knowledge between the source domain and the target domain. The extracted features are then weighted by an attention module to suppress non-transferable local features while enhancing transferable local features. In order to ensure that the data from the target domain in different sub-networks in the same batch is exactly the same, we designed a multi-source domain independent strategy to provide the possibility for later local feature fusion to complete the key features required. In order to verify the effectiveness of the method, we made the dataset "Local Carvana Image Masking Dataset". Applying the proposed method to the image segmentation task of the proposed dataset achieves better transfer performance than other multi-source transfer learning methods. It is shown that the designed transfer learning method is feasible for transfer scenarios where the source and target domains have only local similarities. 
\end{abstract}	

\end{frontmatter}

% \linenumbers

\section{INTRODUCTION}
\label{sec:introduction}
Transfer learning\cite{weiss2016survey} has attracted widespread attention\cite{pan2010survey} due to its ability to reduce the dependence on data volume\cite{tan2018survey,zhao2017research} and computing power\cite{zhang2021federated,zhang2022lightweight}.  A large number of previous works have proved that\cite{cao2010adaptive,dwivedi2019representation} the selection of the source domain is very important, because transfer learning requires that the source domain and the target domain are sufficiently similar to ensure performance. 
For this reason, some transfer learning works empirically consider whether the physical meaning relationship between reality and the target domain is common. For example, the transfer of the same type of objects from the game scene to the real scene\cite{kim2017end}, or the transfer of a certain tissue/organ under different imaging devices, such as ultrasound and CT\cite{RN76}. These are the transfer of things with the same semantic information in different scenarios. In addition, there is a relatively common source domain selection evaluation method that calculates the similarity between two domain data sets through similarity measures and other methods\cite{fawaz2018transfer}. Limited by people's empirical knowledge and similarity measurement methods for datasets, the selected source domain datasets are generally similar to the target domain datasets as a whole.
\par
But what strategy should be adopted when no overall similar source domain can be found? Can some source domain that is locally similar to the target domain be used instead? This decision is difficult, the source and target are different objects, how can one explain why different objects can be chosen as source domains? In addition, if the similarity is calculated quantitatively, only two domains with similar local features may not be able to obtain a high enough similarity evaluation. In short, the source domain, which is similar to the local feature level of the target domain, can only provide limited local features, and it may be difficult to cover all the features required by the target domain. If the non-transferable features are forcibly matched, it may lead to negative transfer\cite{RN9}.
However, we believe that this does not mean that transfer learning cannot be performed, because natural fields have more or fewer similarities such as image contours or textures at the level of local features\cite{RN53}, which means that it is possible to fully mine local features. Although source domains with similar local features can only mine very little information, if the missing information can be filled, even some source domains with similar local features that are very different at the overall level may achieve better results. It is a potential method to extract the transferable knowledge of multiple source domains with similar local features to the target domain in different ways, and then fuse them to complete the knowledge required by the target domain. Even if a single source domain cannot mine enough knowledge, it may be supplemented by other source domains. Therefore, for scenarios where the source domains that are sufficiently similar at the overall level cannot be found, it is possible to achieve the performance of globally similar single-source domain transfer learning by using multi-source transfer learning with similar source domains in local features.
\par
However, we believe that most of the current multi-source transfer learning work only focuses on two problems: How to make the model perform well in different target domains? If single source domain knowledge is lacking, how to obtain sufficient knowledge from other source domains?
\par
Multi-source transfer learning on how to make the model have good performance in different target domains is called domain generalization\cite{RN37,RN26}, and is generally used in scenarios where new target domains are often encountered. For example, for multiple different patients, extract the common features of their ECG or abnormal blood glucose changes to train the model. When encountering a new patient, it can use less data to predict the changes in the patient's physiological indicators\cite{RN49}. There is also work \cite{wan2022uav} that uses data collected from multiple different source areas to pre-train the model, so that the trained model has good performance after fine-tuning in different target areas. The purpose is to extract the most common features in multiple source domains, emphasizing the generalization ability of the model on different data sets, and the idea is similar to taking the intersection of multiple sets. In addition to the idea of domain generalization, the transfer learning of multi-model integration class can also deal with the problem of model generalization in different target domains. Its idea is to use different source domains to independently train multiple models, and then use different models for processing according to the specific conditions of the target domain. For example, through the similarity measurement of each source domain and target domain or according to the performance of the task, the optimal model\cite{RN27,RN29,RN28} is selected for the task. There are also some works that determine the final output results by voting\cite{she2022multi} for all model outputs when facing classification problems; some works\cite{RN41,RN24} also design different weights for different models. The idea of multi-model integration is similar to that of multi-sensor measurement to improve accuracy, the more models there are, the greater the possibility of good performance.
\par
Regarding the lack of data in a single source domain, the simplest idea is to mix data from all source domains into one new source domain\cite{RN35,sun2015survey}. This is generally applicable to scenarios where the amount of data in the source domain is insufficient and each source domain has a large similarity with the target domain. Some works\cite{RN45,RN33,RN30,RN32,RN36,RN65} also assign different weights to samples from different source domains according to the overall similarity between source and target domains. There are also works that select the most suitable samples\cite{RN67} in multiple source domains to augment data, such as designing classifiers\cite{RN43}. Besides, there are some works that transform the labels\cite{RN21} or samples\cite{yao2010boosting} of the target domain and multiple source domains into the same space, and then treat them as one source domain in the new space. There are also some works\cite{RN50} that try to explore in terms of internal features of deep networks. Their idea is that different middle and high-level features can be composed by sharing low-level features. This type of work extracts low-level features through the same extractor, and the low-level features are then aligned with mid-level and high-level feature extractors designed for each source, and finally complete the target domain task. Specifically, these works feed data from all domains into the same network front-end layer, in an attempt to find common low-level features that can be represented by each domain. Of course, the sharing of low-level features of deep networks has been demonstrated by Yosinski et al.\cite{RN53}.
\par
Based on the previous work, we believe that current multi-source transfer learning work still only focuses on the overall similarity between source and target domains. In addition to the ideological level, many works at the experimental level still use the dataset with consistent semantic information in the source domain and the target domain as the standard dataset for experimental verification. For example the public Office-Caltech10 object dataset\cite{gong2012geodesic,griffin2007caltech}. There are also works such as\cite{zuo2021attention} using MNIST\cite{lecun1998gradient}, MNIST-M\cite{ganin2015unsupervised}, USPS\cite{hull1994database}, SVHN\cite{netzer2011reading} and Synthetic Digits\cite{ganin2015unsupervised} multiple handwritten digit datasets.
\par
These works cannot perform multi-source transfer learning based on local features similar to source domains. To this end, we hope to redefine a multi-source transfer learning strategy at the local feature level to solve the problem of multiple source transfer learning scenarios with only local similarity to the target domain. Different from these works, our work pays more attention to the complementary local features between each source domain. In the face of a scene where there is no global similarity to the source domain, how to mine local common features from multiple source domains with limited global similarity and integrate them to improve the task performance of the target domain is a problem we need to solve. Different from the existing multi-source transfer learning work, the idea of this work is to use multiple source domains to complete the lack of local features. Using multiple source domains that have different local features similar to the target domain, by mining transferable local features for each source domain, they are finally fused for transfer learning. Taking the deep network as an example, in the process of network learning from end to end, there are actually many deep features that have the potential to be invariants\cite{RN53}. Because the deep features of these black boxes are uninterpretable, it is difficult to manually design rules to mine transferable features of each source domain and fuse them. We believe this is the reason why the current multi-source deep transfer learning source domain selection is limited to the overall similarity of the target domain. If it can be determined which deep features are universal, it is possible to relax the selection of source domains, so that multi-source transfer learning is no longer limited to selecting source domains with similar overall features, and only needs to select similar local features. In terms of local feature selection and enhancement, the attention mechanism\cite{vaswani2017attention} is derived from Transformer, which can adaptively enhance transferable feature representations and suppress unnecessary feature representations. Adopting attention mechanism to filter relevant features in each source domain is a way to solve this problem. As we all know, the human attention mechanism not only pays attention to the whole object, but also pays more attention to the local useful information of the object. Inspired by the human attention mechanism, some works that introduce the attention mechanism make it possible to mine and enhance the local features of objects, so that important features can be distinguished from insignificant features. For example\cite{long2018attention} apply separate attention clusters to different feature sets and then concatenate the outputs for classification. We argue that the idea of distinguishing whether local features are important or not is necessary for domain adaptation, especially when each source domain has only some local feature similarities. Because the transferability of different local features is different for the same domain, we should distinguish transferable and non-transferable local features, such as assigning greater weight to important local features.
\par
In response, we propose an attention mechanism-based Multi-source Adversarial Transfer Learning approach to handle transfer scenarios with only local feature-level similar source domains. The proposed method extracts common local features of each source and target domain separately through multiple sub-networks. Specifically, the sub-network feature extractor extracts transferable local features through domain classifier adversarial training. The attention module weights the features extracted by the feature extractor, giving greater weight to transferable local features, while suppressing non-transferable local features. Input the fused local features into the predictor to get the prediction result. In addition, the method also designs a multi-source domain independent strategy, so that the same batch of target domain data entering different sub-networks is exactly the same. During the training process, each data from the source domain can complete confrontation training with the same target domain data, thereby ensuring the feasibility of the fusion of the extracted general features. Since the existing multi-source transfer learning labeled datasets are used to verify the overall similarity-based multi-source transfer learning algorithm, they are not available in this work. To validate the proposed method, we propose the dataset "\href{https://github.com/ZhangYixiaofu/Local-Carvana-Image-Masking-Dataset}{Local Carvana Image Masking Dataset}" for validating multi-source transfer learning based on similar source domains of local features. The proposed dataset is based on the “Carvana Image Masking Challenge” dataset released by Shaler et al. on kaggle\cite{carvana}. The proposed method is applied to image segmentation tasks, and achieves better transfer performance than other similar methods, indicating that the method has good applicability in transfer scenarios with similar local features. Our work has the following contributions:
\\1. For the scenario where the source domain and the target domain only have similar local features, a multi-source transfer learning method based on local feature similarity to the source domain is designed.
\\2. In the scenario where the source domain and the target domain only have similar local features, the extracted features may have the problem of forcibly matching the non-transferable features, and the self-attention mechanism is used to screen the transferable features.
\\3. For the fusion of local features extracted from multiple source domains, a multi-source domain independent strategy is proposed to ensure that each sub-network uses the same target domain data during the training process, thereby ensuring the fusion of local features.
\par
The rest of this paper is organized as follows. Section \ref{METHODS} presents the proposed Multi-source Adversarial Transfer Learning based on similar source domains with local features. Section \ref{EXPERIMENTAL} verifies the transfer performance of the proposed method on the released dataset "Local Carvana Image Masking Dataset". Section \ref{CONCLUSION} concludes the whole work.

\section{METHODS}\label{METHODS}
\subsection{Problem definition}
Given a target domain dataset ${\cal D}_t^{} = \left\{ {x_j^{},y_j^{}} \right\}_{j = 1}^{{n_{{t}}}}$, 
among them, ${x_j}$ is the target domain sample, ${y_j}$ is the target domain label.
At the same time, a set of source domains ${\cal D}_s^{} = \left\{ {{{\cal D}_{{s_i}}}} \right\}_{i = 1}^N$ with $N$ source domains is given, and each source domain has ${n_{{s_i}}}$ data with labels, denoted as ${{\cal D}_{{s_i}}} = \left\{ {x_j^i,y_j^i} \right\}_{j = 1}^{{n_{{s_i}}}}$, where $x_j^i$ is the source domain data and 
$y_j^i$ is the source domain label. The source and target domains come from domains with similar local features.
\par
The considered task is to fuse and transfer knowledge from $N$ source domain ${{\cal D}_{{s_i}}} = \left\{ {x_j^i,y_j^i} \right\}_{j = 1}^{{n_{{s_i}}}}$ with similar local features to target domain ${\cal D}_t^{} = \left\{ {x_j^{},y_j^{}} \right\}_{j = 1}^{{n_{{t}}}}$ to improve the task effect. Since the respective source domains and target domain only have local feature similarities, their marginal probabilities present different distributions: $P\left( {{{\cal D}_t}} \right) \ne P\left( {{{\cal D}_{{s_1}}}} \right) \ne  \cdots  \ne P\left( {{{\cal D}_{{s_i}}}} \right) \ne  \cdots  \ne P\left( {{{\cal D}_{{s_n}}}} \right)$. Therefore, the goal of multi-source transfer learning in this paper is to learn multiple transformation functions $F_T^i\left(  \cdot  \right)\left| {_{i = 1}^N} \right.$, to let $\left\{ {\begin{array}{*{20}{c}}
		{P\left( {F_T^i\left( {{{\cal D}_t}} \right)} \right) = P\left( {F_T^i\left( {{{\cal D}_{{s_i}}}} \right)} \right)}\\
		{\sum\limits_{i = 1}^N {F_T^i\left( {{{\cal D}_t}} \right)}  = \sum\limits_{i = 1}^N {F_T^i\left( {{{\cal D}_{{s_i}}}} \right)} }
\end{array}} \right.$. When performing the task, this function will be used as a part of the prediction function ${{\cal F}_P}\left(  \cdot  \right)$ to complete the domain adaptation work of each source domain and target domain in the form of ${{\cal F}_P}\left( {\sum\limits_{i = 1}^N {F_T^i\left(  \cdot  \right)} } \right)$.
\subsection{The overall idea of Multi-source Adversarial Transfer Learning based on local feature similar source domain}
The overall idea of Multi-source Adversarial Transfer Learning based on similar source domains of local features is to use a multi-source domain independent strategy in the data processing stage to obtain multiple sub-batch data, and then input each sub-batch data into the corresponding sub-network. After each sub-network extracts the common local features of the corresponding source domain and target domain, it enters the fusion predictor to obtain the prediction result. As shown in Figure \ref{fig_overall}, our approach consists of three categories of modules, including:
\begin{figure}[!t]
	\centerline{\includegraphics[width=\columnwidth]{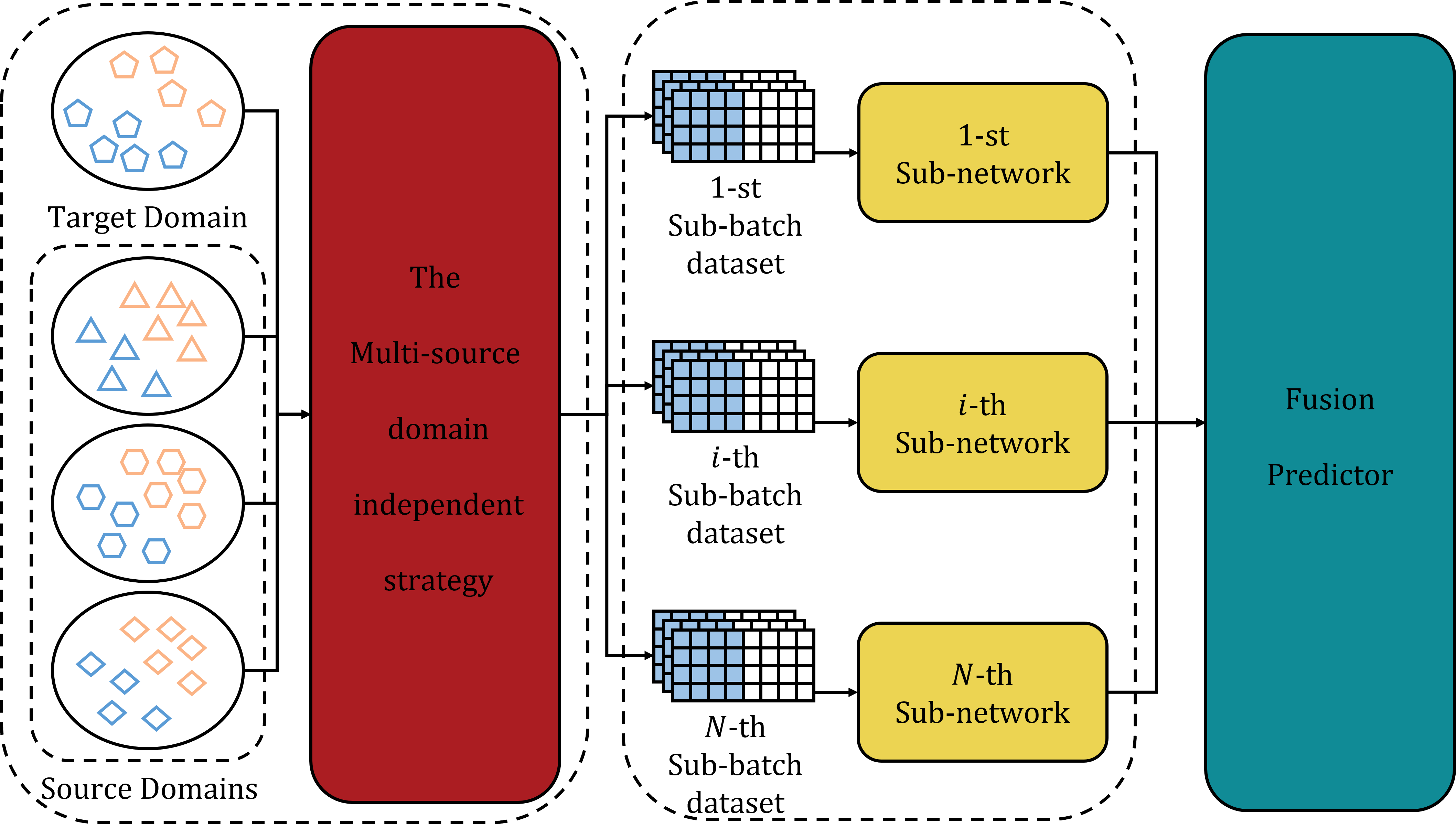}}
	\caption{The overall idea of Multi-source Adversarial Transfer Learning based on local feature similar source domain.}
	\label{fig_overall}
\end{figure}
\\(1) Multi-source domain independent strategy. After the multi-source domain independent strategy, $N$ source domains are divided into $N$ types of sub-batch data together with the target domain to ensure the possibility of later fusion. See section \ref{independent} for details.
\\(2) Adversarial transfer learning sub-network for extracting local features, $N$ sub-batch data sets enter $N$ sub-networks, and extract local features that can be transferred between source domain-target domain combinations. They are then given greater weight while suppressing non-transferable local features. See Section \ref{Subnetwork} for details.
\\(3) Predictor, $N$ The local features from the sub-network are input into the predictor, and then the prediction result is obtained. It is similar to the overall design of Multi-source Adversarial Transfer Learning based on local features, see Section \ref{overall}.
\subsection{The Multi-source domain independent strategy}\label{independent}
\begin{figure}[!t]
	\centerline{\includegraphics[width=0.6\columnwidth]{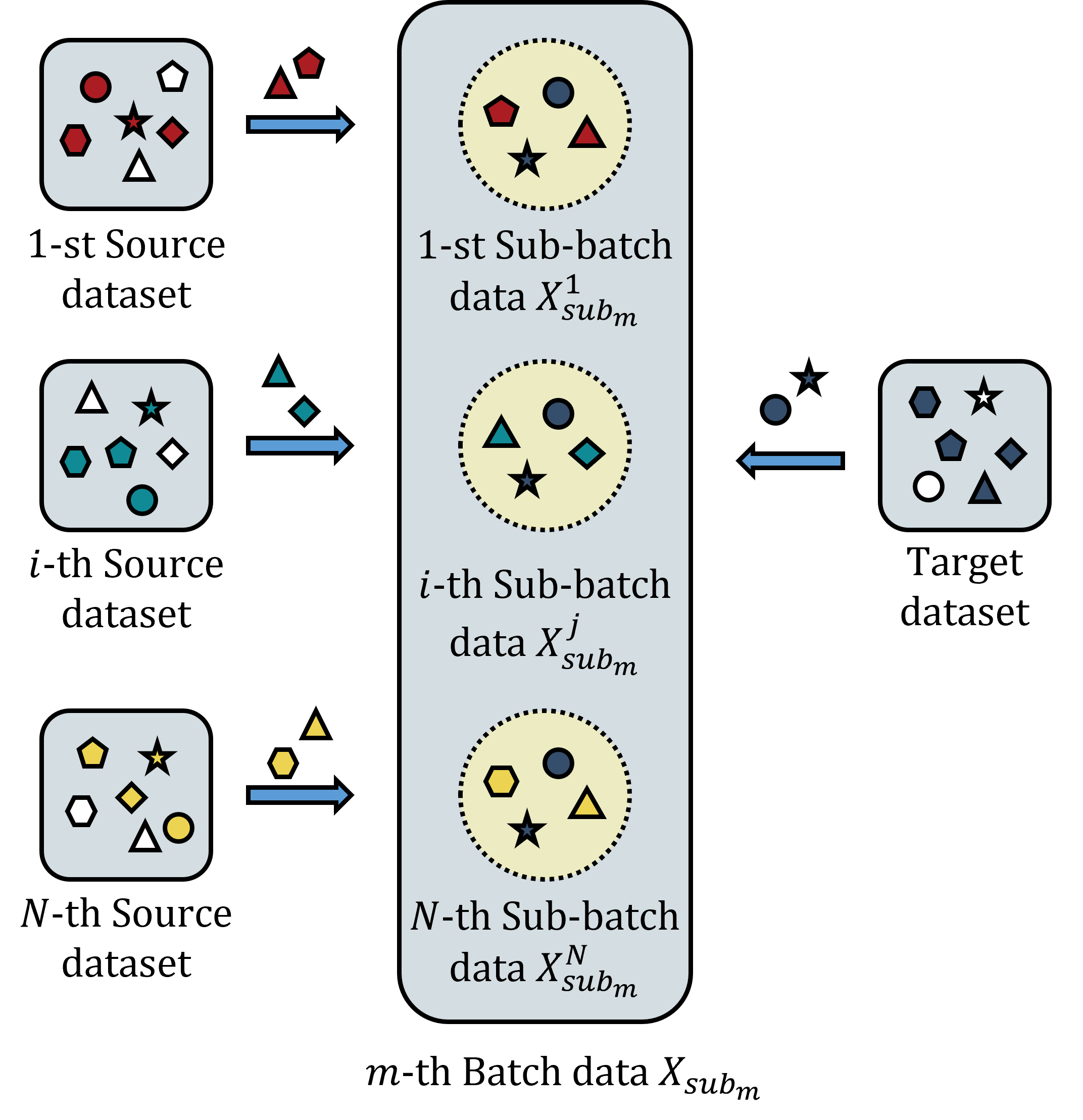}}
	\caption{Multi-source domain independent strategy, the data from the target domain will form $N$ sub-batches with the same amount of data from $N$ source domains, and the source domain data of each sub-batch is the same. $N$ sub-batches form a batch, and each sub-batch corresponds to a sub-network when training the network.}
	\label{fig_b}
\end{figure}

Because this work designs an object predictor for fusing local features from multiple source domains, it is necessary to ensure the fusion ability of local features provided by each source domain. In Section 2.5, we use the sum method for fusion, so it is necessary to ensure that the target domain data passing through each sub-network in each batch is the same. The domain independence strategy \cite{RN69} is a strategy we proposed in the past to ensure data balance. Its idea is to select the same amount of data to form batch data when different domains remain independent. We extend the domain independence strategy to multiple source domains to ensure that the target domain data across each sub-network within each batch is the same. The core idea of the extension is to select the same amount of data in each domain while the target domain remains independent from all source domains. The exact same target domain data and data from different source domains form Sub-batch data, and then enter the sub-networks respectively to complete the confrontation transfer learning.Specifically, as shown in Figure \ref{fig_b}, suppose a batch has a total of $n_b^{}$ data, including $N$ sub-batches in which individual source domain data and target domain data are mixed, and each sub-batch has ${n_{sub - b}} = \frac{{{n_b}}}{N}$ samples. 
The multi-source domain independent strategy first takes $\frac{{{n_{sub - b}}}}{2} = \frac{{{n_b}}}{{2 \times N}}$ data samples in the target domain, and then also takes $\frac{{{n_{sub - b}}}}{2}$ data samples in each source domain. Then the samples drawn from each source domain and the samples drawn from the target domain are mixed into a sub-batch, and domain labels are added to them at the same time. $N$ source domains will be mixed into $N$ sub-batches, and the data from the target domain in each sub-batch is exactly the same. Repeat $batc{h_{total}} = \frac{{\max \left( {\left( {{n_{{s_i}}}} \right)\left| {_{i = 1}^N} \right.} \right)}}{{{{{n_{sub - b}}} \mathord{\left/
				{\vphantom {{{n_{sub - b}}} 2}} \right.
				\kern-\nulldelimiterspace} 2}}}$ times until all the data of each source domain has entered the Sub-batch. See Algorithm \ref{alg1} for the specific process.
\begin{algorithm}[H]
	\caption{Multi-source domain independent strategy}
	\label{alg1} % 引用标签
	\begin{algorithmic}[1]	 
		\REQUIRE ~~\\ %算法的输入参数：Input
		The source datasets, ${\cal D}_s^{} = \left\{ {{{\cal D}_{{s_i}}}} \right\}_{i = 1}^N$, Where, ${{\cal D}_{{s_i}}} = \left\{ {x_j^i,y_j^i} \right\}_{j = 1}^{{n_{{s_i}}}}$;\;
		The target dataset, $D_t^{}$, Where, $D_t^{} = \left\{ {x_j^{},y_j^{}} \right\}_{j = 1}^{{n_{{t}}}}$;
		\FOR{$m = 1; m \le \frac{{\max \left( {\left( {{n_{{s_i}}}} \right)\left| {_{i = 1}^N} \right.} \right)}}{{{{{n_{sub - b}}} \mathord{\left/
							{\vphantom {{{n_{sub - b}}} 2}} \right.
							\kern-\nulldelimiterspace} 2}}}; m +  + $}    
		\STATE Select $\frac{{{n_{sub - b}}}}{2} = \frac{{{n_b}}}{{2 \times N}}$ samples from the target dataset ${\cal D}_t^{}$ as the target sub-batch dataset $X_{su{b_m}}^t$;\;
		\FOR{$i = 1; i \le N; i +  + $}
		\STATE Select $\frac{{{n_{sub - b}}}}{2} = \frac{{{n_b}}}{{2 \times N}}$ samples from the source dataset ${{\cal D}_{{s_i}}}$ as the $i-th$ source sub-batch dataset 	$X_{su{b_m}}^{{s_i}}$;\;
		\STATE Mix the target sub-batch dataset $X_{su{b_m}}^t$ and the $i-th$ source sub-batch dataset $X_{su{b_m}}^{{s_i}}$ as sub-batch dataset $X_{su{b_m}}^i$;
		\ENDFOR
		\ENDFOR
		\ENSURE ~~\\ %算法的输出：Output
		Sub-batch datasets,	$X_{su{b_{}}}^{} = \left\{ {\left\{ {X_{su{b_m}}^i} \right\}_{i = 1}^N} \right\}_{m = 1}^\frac{{\max \left( {\left( {{n_{{s_i}}}} \right)\left| {_{i = 1}^N} \right.} \right)}}{{{{{n_{sub - b}}} \mathord{\left/
				{\vphantom {{{n_{sub - b}}} 2}} \right.
				\kern-\nulldelimiterspace} 2}}}$ ; 
	\end{algorithmic}
\end{algorithm}

\subsection{The Adversarial Transfer Learning Sub-network for Extracting Local Features}\label{Subnetwork}
The adversarial transfer learning network\cite{ganin2015unsupervised} is inspired by the generative adversarial network\cite{goodfellow2014generative}, and its structure is generally composed of a feature extractor, an output predictor and a domain classifier. Its core idea is to adaptively extract common features of source and target domains through the confrontation of feature extractor and domain discriminator. Then the common feature is used as the input of the target predictor to achieve the goal of task performance optimization in the target domain, and also provides the possibility to utilize the unlabeled target domain data to the greatest extent.
For the $i-th$ sub-network, the $i-th$ feature extractor $G_e^i\left( { \cdot ;\theta _e^i} \right)$ is used to extract the common features ${\cal X}_{}^i$ of the $i-th$ source and target domains. The features extracted by the source domain data $x_j^i$ through the feature extractor are denoted as ${{\cal X}}_{}^{{s_i}} = G_e^i\left( {x_j^i;\theta _e^i} \right)$. The features extracted by the target domain data $x_j^{}$ through the feature extractor are denoted as ${\cal X}_{}^t = G_e^i\left( {x_j^{};\theta _e^i} \right)$. An extracted feature is considered transferable if the domain classifier cannot tell which domain the feature comes from.
\par 
Since the data input by the network in this work may only be transferable at the level of local features, it is very important to accurately screen transferable local features. Because if the feature extractor extracts features that are not transferable, it will cause negative transfer. Inappropriate deep features may even cause significant errors in the training process of domain discriminators and predictors, and these errors will also be back-propagated to feature extractors and further amplified. To this end, inspired by \cite{RN9} and \cite{zhang2019self}, an attention mechanism is introduced to adaptively enhance transferable features and suppress non-transferable features. However, due to the limited integrity of the source domain and the target domain, there may be fewer transferable features, which will lead to the problem of insufficient key features. In the next section, we will discuss how to use multiple source domains with similar local features to complete.
\begin{figure}[!t]
	\centerline{\includegraphics[width=\columnwidth]{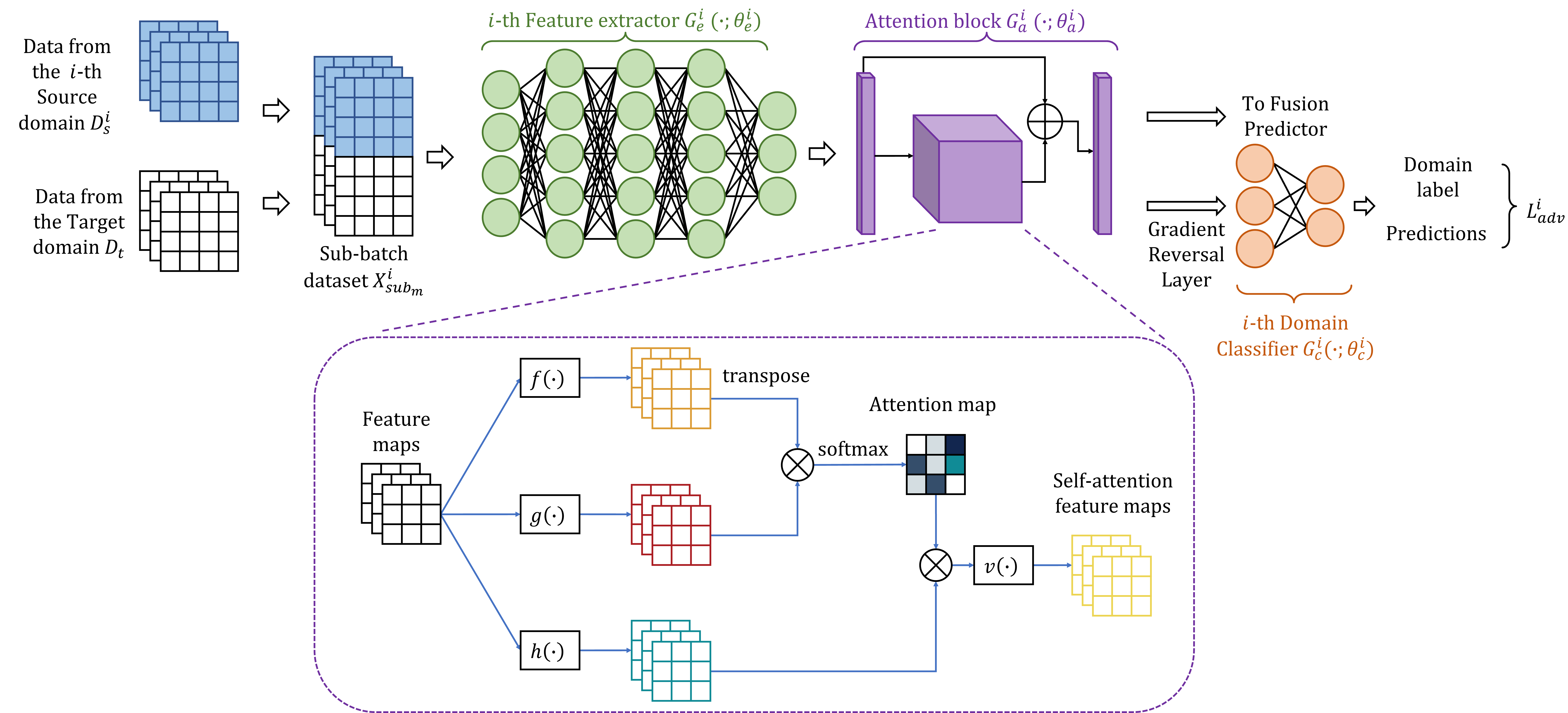}}
	\caption{Sub-network, the feature extractor is encouraged to extract common features between the source and target domains through the domain classifier, and the transferable local features are screened through the attention mechanism.}
	\label{fig_3}
\end{figure}
The sub-network structure is shown in Figure \ref{fig_3}. The self-attention module inputs the feature ${\cal X}_{}^i \in R_{}^{C \times H \times W}$ extracted by the feature extractor $G_e^i\left( { \cdot ;\theta _e^i} \right)$, and then reshape the matrix of size $R_{}^{C \times K}$, where, $K = H \times W$, to perform matrix operations. Among them, $C$ is the number of channels, $H$ is the height of the feature map, and $W$ is the width of the feature map. In order to calculate attention, ${\cal X}_{}^i$ first enters two feature spaces $f$ and $g$, and obtains two abstract feature maps $f\left( {{\cal X}_{}^i} \right) = W_f^i{\cal X}_{}^i$ and $g\left( {{\cal X}_{}^i} \right) = W_g^i{\cal X}_{}^i$, then uses softmax operation, and calculates the self-attention map ${M_{att}^i} = \left[ {\begin{array}{*{20}{c}}
		{m_{1,1}^{i}}&{m_{1,2}^{i}}& \cdots &{m_{1,K}^{i}}\\
		{m_{2,1}^{i}}&{m_{2,2}^{i}}& \cdots &{m_{2,K}^{i}}\\
		{m_{3,1}^{i}}&{m_{3,2}^{i}}& \cdots &{m_{3,K}^{i}}\\
		\vdots & \vdots &{m_{b,a}^{i}}& \vdots \\
		{m_{K,1}^{i}}&{m_{K,2}^{i}}& \cdots &{m_{K,K}^{i}}
\end{array}} \right]$ with Eq.\ref{eq1},
where $m_{b,a}^{i}$ indicates how much the model pays attention to the $b-th$ location when synthesizing the $a-th$ region, the more similar two location features are, the greater the correlation between them. $W_f^i$ and $W_g^i$ are parameters that need to be learned.
\begin{equation}
	m_{b,a}^{i} = \frac{{\exp \left( {f\left( {{\cal X}_a^i} \right)_{}^Tg\left( {{\cal X}_b^i} \right)} \right)}}{{\sum\limits_{a = 1}^K {\exp \left( {\left( {f\left( {{\cal X}_a^i} \right)_{}^Tg\left( {{\cal X}_b^i} \right)} \right)} \right)} }}\label{eq1}
\end{equation}
\\
At the same time, the abstract feature map ${\cal X}_{}^i$ reshaped into $R_{}^{C \times K}$ enters the feature space to obtain the abstract feature map $h\left( {{\cal X}_{}^i} \right) = W_h^i{\cal X}_{}^i$, and then $h\left( {{\cal X}_{}^i} \right) = W_h^i{\cal X}_{}^i$ and the transposed ${M_{att}^i}$ enter the feature space $v$ and calculate ${{\cal X}}_b^{i^{'}}$ according to Eq.\ref{eq2},
\begin{equation}
	{{\cal X}}_b^{i^{'}} = v\left( {\sum\limits_{a = 1}^K {m_{b,a}^{i}h\left( {{\cal X}_{}^i} \right)} } \right) = W_v^i\left( {\sum\limits_{a = 1}^K {m_{b,a}^{i}h\left( {{\cal X}_{}^i} \right)} } \right)\label{eq2}
\end{equation}
The above $W_h^i$ and $W_v^i$ are the parameters that need to be learned. After calculating all the X results, the output Self-attention feature map ${{\cal X}}^{i^{'}}$ can be obtained as Eq.\ref{eq3}.
\begin{equation}
{{\cal X}}^{i^{'}} = \left( {\begin{array}{*{20}{c}}
		{{{\cal X}}_1^{i^{'}},}&{{{\cal X}}_2^{i^{'}},}& \cdots &{{{\cal X}}_b^{i^{'}},}& \cdots &{{{\cal X}}_K^{i^{'}}}
\end{array}} \right)\label{eq3}
\end{equation}
We multiply the output ${{\cal X}}^{i^{'}}$ of the attention layer by a scale parameter $\mu$ and add back the input feature map ${\cal X}_{}^i$. Therefore, the final output ${\cal X}_{att}^i$ is:
\begin{equation}
{\cal X}_{att}^i = \mu {{\cal X}}^{i^{'}} + {\cal X}_{}^i\label{eq4}
\end{equation}
Among them, the scalar $\mu  \ge 0$, according to the research of Chen et al.\cite{chen2019transferability}, in general, $\mu  = 0$ will be initialized, and will be assigned a greater weight as the training progresses. Denote the parameters of the attention module as $\theta _a^i$:
\begin{equation}
\theta _a^i = \left\{ {W_f^i;W_g^i;W_h^i;W_v^i} \right\}\label{eq5}
\end{equation}
After the estimated parameter $\widehat {\theta _a^i}$ of the attention module is obtained through training, the abstract feature map ${\cal X}_{att}^i$ available after weighting can be obtained.
\subsection{The overall design of Multi-source Adversarial Transfer Learning based on local feature similar source domain}\label{overall}
\begin{figure}[!t]
	\centerline{\includegraphics[width=\columnwidth]{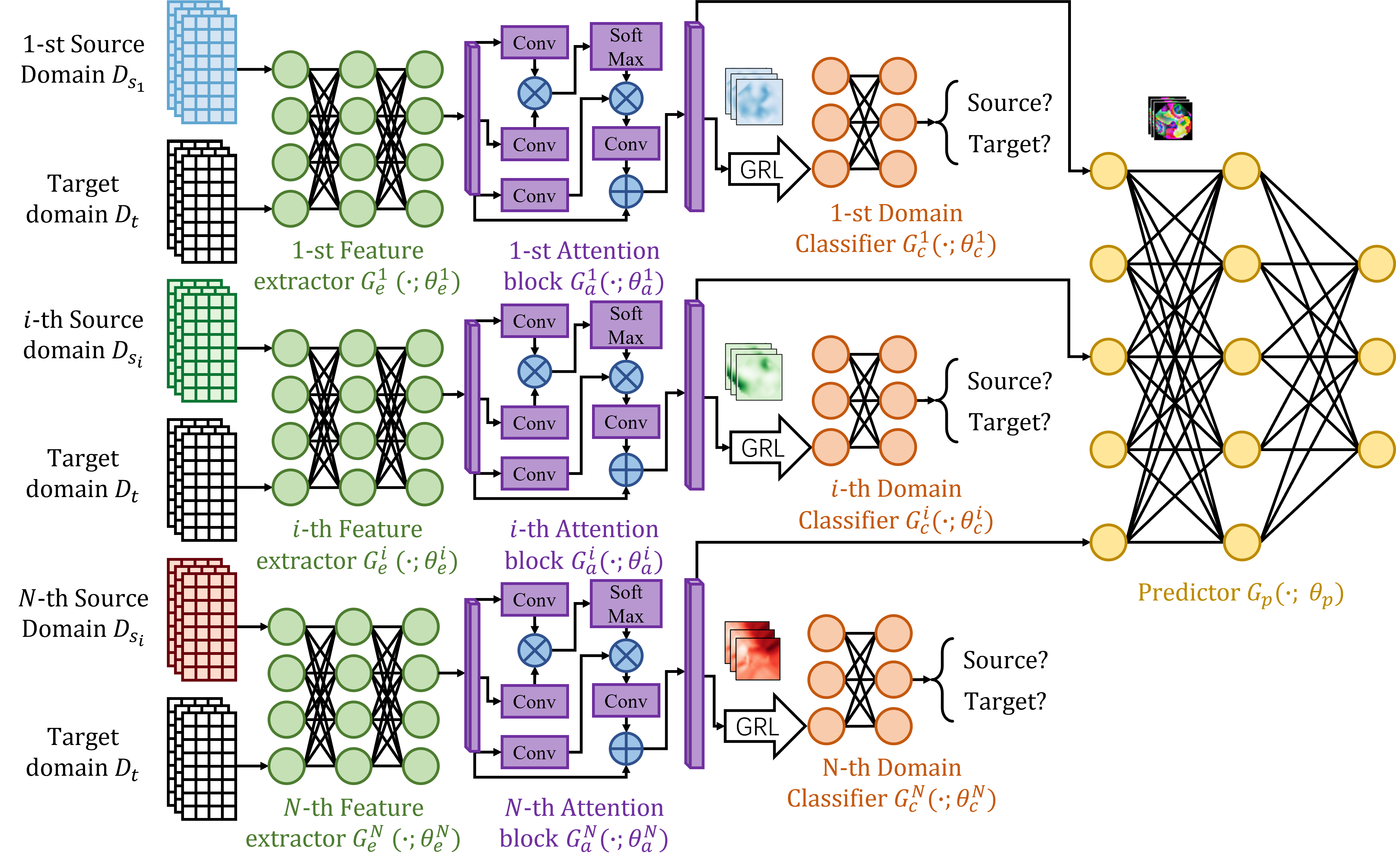}}
	\caption{Overall network, Multiple sub-networks extract and screen transferable local features, which are fused in the predictor by summation. The predictor outputs prediction results.}
	\label{fig_4}
\end{figure}

The purpose of the proposed Multi-source Adversarial Transfer Learning is to learn the knowledge of the source domain set ${\cal D}_s^{} = \left\{ {{{\cal D}_{{s_i}}}} \right\}_{i = 1}^N$ to assist the improvement of the target domain task performance. Specifically, multiple sub-networks should extract the general local features of each source domain ${{\cal D}_{{s_i}}}$ and target domain ${\cal D}_t^{}$. As the number of learning iterations increases, the self-attention mechanism module of each sub-network will give higher weight to the features with higher correlation, and gradually emphasize the local features that can be transferred. Finally, all the required deep features are completed through information fusion.
Due to the designed multi-source domain-independent strategy, in the $m-th$ sub-batch data $X_{su{b_m}}^i$ passing through the $i-th$ sub-network, the data from the target domain $X_{su{b_m}}^t$ is exactly the same. Therefore, the data of the same batch of different sub-networks used to extract the characteristics of the target domain is completely consistent. In addition, as shown in Figure \ref{fig_4}, the structure of the sub-networks is also exactly the same, so they can be fused by summing as $\sum\limits_{i = 1}^N {{\cal X}_{att}^i} $.
\par
The sub-batch data are used to train multiple feature extractors $G_e^i\left( { \cdot ;\theta _e^i} \right)$ to obtain local features. The transferable local features are then augmented by an attention module $G_a^i\left( { \cdot ;\theta _a^i} \right)$. The domain discriminator of the sub-network is used to extract general features. The more general features are less likely to be identified from which domain, the optimization goal should be Eq.\ref{eq6}:
\begin{equation}
\mathop {\arg }\limits_{\max } \left( {L_{adv}^i} \right) = \mathop {\arg }\limits_{\max } L_{adv}^i\left( {G_c^i\left( {G_a^i\left( {G_e^i\left( { \cdot ;\theta _e^i} \right);\theta _a^i} \right);\theta _c^i} \right)} \right)\label{eq6}
\end{equation}
Since multiple source domains are involved, the deep features output by each feature extraction sub-network need to be fused, and the summation strategy is used for local feature fusion. 
Specifically, in terms of the loss function of the target domain and the $i-th$ source domain $S_i$ for adversarial transfer learning, the target domain task requires the prediction performance to be as good as possible, and the optimization goal should be:
\begin{equation}
\mathop {\arg }\limits_{\min } \left( {L_p^{}} \right) = \mathop {\arg }\limits_{\min } \left( {L_p^{}\left( {{G_p}\left( {\sum\limits_{i = 1}^N {G_a^i\left( {G_e^i\left( { \cdot ;\theta _e^i} \right);\theta _a^i} \right)} ;{\theta _p}} \right)} \right)} \right)\label{eq7}
\end{equation}
Therefore, the overall optimization goal is:
\begin{equation}
\begin{array}{l}
	{\cal L}\left( {{X_j}} \right) = L_p - \lambda \sum\limits_{i = 1}^N L_{adv}^i = L_p^{}\left( {{G_p}\left( {\sum\limits_{i = 1}^N {G_a^i\left( {G_e^i\left( {{X_j};\theta _e^i} \right);\theta _a^i} \right)} ;{\theta _p}} \right)} \right)\\
	- \lambda \sum\limits_{i = 1}^N {L_{adv}^i\left( {G_c^i\left( {G_a^i\left( {G_e^i\left( {{X_j};\theta _e^i} \right);\theta _a^i} \right);\theta _c^i} \right)} \right)} \label{eq8}
\end{array}
\end{equation}
where $\lambda$ is the balance parameter.
In terms of parameter optimization goals, the data from the source domain and the target domain will be given domain labels, and input the domain classifier $G_c^i\left( {G_e^i\left( {{X_j};\theta _e^i} \right);\theta _c^i} \right)$ will complete the adversarial domain adaptation. Then the local common features obtained in the sub-networks will be fused by summing and used as the input of the predictor ${G_p}\left( {\sum\limits_{i = 1}^N {G_a^i\left( {G_e^i\left( {{X_j};\theta _e^i} \right);\theta _a^i} \right)} ;{\theta _p}} \right)$, and the outputs of multiple sub-networks will be fused to form the output. Therefore, the optimization objective of the parameters is:
\begin{equation}
\left\{ {\widehat {{\theta _p}};\left\{ {\widehat {\theta _e^i};\widehat {\theta _a^i};\widehat {\theta _c^i}} \right\}_{i = 1}^N} \right\} = \mathop {\min }\limits_\Theta  \left( {{\cal L}\left( \Theta  \right)} \right)\label{eq9}
\end{equation}
Where $\Theta  = \left\{ {{\theta _p};\left\{ {\theta _e^i;\theta _a^i;\theta _c^i} \right\}_{i = 1}^N} \right\}$ is a network parameter and $\hat \Theta  = \left\{ {\widehat {{\theta _p}};\left\{ {\widehat {\theta _e^i};\widehat {\theta _a^i};\widehat {\theta _c^i}} \right\}_{i = 1}^N} \right\}$ is a network estimated parameter. After the network is trained, the learned transformation function is $F_T^i = G_a^i\left( {G_e^i\left( { \cdot ;\theta _e^i} \right);\theta _a^i} \right)$, which can extract common local features between the source domain and the target domain. The prediction function ${{\cal F}_P} = {G_p}\left( { \cdot ;{\theta _p}} \right)$ utilizes the learned common features and has good generalization ability. Finally, the trained prediction function can be directly applied to the prediction task of the target domain.
\begin{equation}
\widehat {{{\cal F}_P}}\left(  \cdot  \right) = {G_p}\left( {\sum\limits_{i = 1}^N {G_a^i\left( {G_a^i\left( { \cdot ;\widehat {\theta _e^i}} \right);\widehat {\theta _a^i}} \right)} ;\widehat {{\theta _p}}} \right)\label{eq10}
\end{equation}
The specific implementation steps are shown in the algorithm \ref{alg:Framwork}
\begin{algorithm}[h]
	%\tiny
	\caption{Multi-source Adversarial Transfer Learning Algorithm} % 标题
	\label{alg:Framwork} % 引用标签
	\begin{algorithmic}[1]	 
		\REQUIRE ~~\\ %算法的输入参数：Input
		The source datasets, ${\cal D}_s^{} = \left\{ {{{\cal D}_{{s_i}}}} \right\}_{i = 1}^N$, Where, ${{\cal D}_{{s_i}}} = \left\{ {x_j^i,y_j^i} \right\}_{j = 1}^{{n_{{s_i}}}}$;
		The target dataset, $D_t^{}$, Where, $D_t^{} = \left\{ {x_j^{},y_j^{}} \right\}_{j = 1}^{{n_{{t}}}}$;
		Total epoch, $e_{total}$;\\
		\STATE Multi-source domain independent strategy as the Algorithmic \ref{alg1};
		\FOR {$e=1$; $e \leq e_{total}$; $e++$}
		\FOR {$m = 1; m \le \frac{{{n_{{s_i}}}}}{{{n_b}}}; m +  + $}
		\FOR{$i = 1; i \le N; i +  + $}
		% \STATE Simultaneously get $b_{s}$ samples from $D^i$ as the batch dataset $D^i_m=\left\{{X^i_m,Y^i_m}\right\}$
		\STATE Extract local features $G_a^i\left( {G_e^i\left( {X_m^i;\theta _e^i} \right);\theta _a^i} \right)$ from the $i-th$ sub-batch dataset $X_m^i$;
		\STATE Get predictions $G_c^i\left( {G_a^i\left( {G_e^i\left( {X_m^i;\theta _e^i} \right);\theta _a^i} \right);\theta _c^i} \right)$;
		\ENDFOR
		\STATE Information fusion $\sum\limits_{i = 1}^N {G_a^i\left( {G_e^i\left( {X_m^i;\theta _e^i} \right);\theta _a^i} \right)} $;\;
		\STATE Get predictions	$G_p^{}\left( {\sum\limits_{i = 1}^N {G_a^i\left( {G_e^i\left( {X_m^i;\theta _e^i} \right);\theta _a^i} \right)} ;\theta _p^{}} \right)$;
		\STATE Calculate the loss function as Eq.\ref{eq8}
		\STATE Backpropagation to update the network parameters $\Theta  = \left\{ {{\theta _p};\left\{ {\theta _e^i;\theta _a^i;\theta _c^i} \right\}_{i = 1}^N} \right\}$ from
		${\cal L}$;
		\ENDFOR
		\ENDFOR
		\ENSURE ~~\\ %算法的输出：Output
		Deep prediction model, $\widehat {{{\cal F}_P}}\left(  \cdot  \right) = {G_p}\left( {\sum\limits_{i = 1}^N {G_a^i\left( {G_e^i\left( { \cdot ;\widehat {\theta _e^i}} \right);\widehat {\theta _a^i}} \right)} ;\widehat {{\theta _p}}} \right)$;
	\end{algorithmic}
\end{algorithm}

\section{EXPERIMENTAL RESULTS AND ANALYSIS}\label{EXPERIMENTAL}
In transfer learning applications, conventional transfer learning is usually not available since the source and target domains only have local feature-level similarities. To solve this problem, fusing the source domain information of multiple local feature level similarities has very important application value. Therefore, this paper uses the data set provided by "Carvana Image Masking Challenge"\cite{carvana}, and uses Gaussian noise to generate two source domain data sets of body and wheels from the cars in the data set. Then the proposed method is verified.
\par
The experimental environment computer is NVIDIA GeForce RTX 3090 GPU, 12th Gen Intel® Core™ i9-12900K Processor, with 32 Gb memory, and the development tools are Python 3.9 and Pytorch 1.10.2.
\subsection{"Carvana Image Masking Challenge" Dataset description}
\begin{figure}[!t]
	\centerline{\includegraphics[width=\columnwidth]{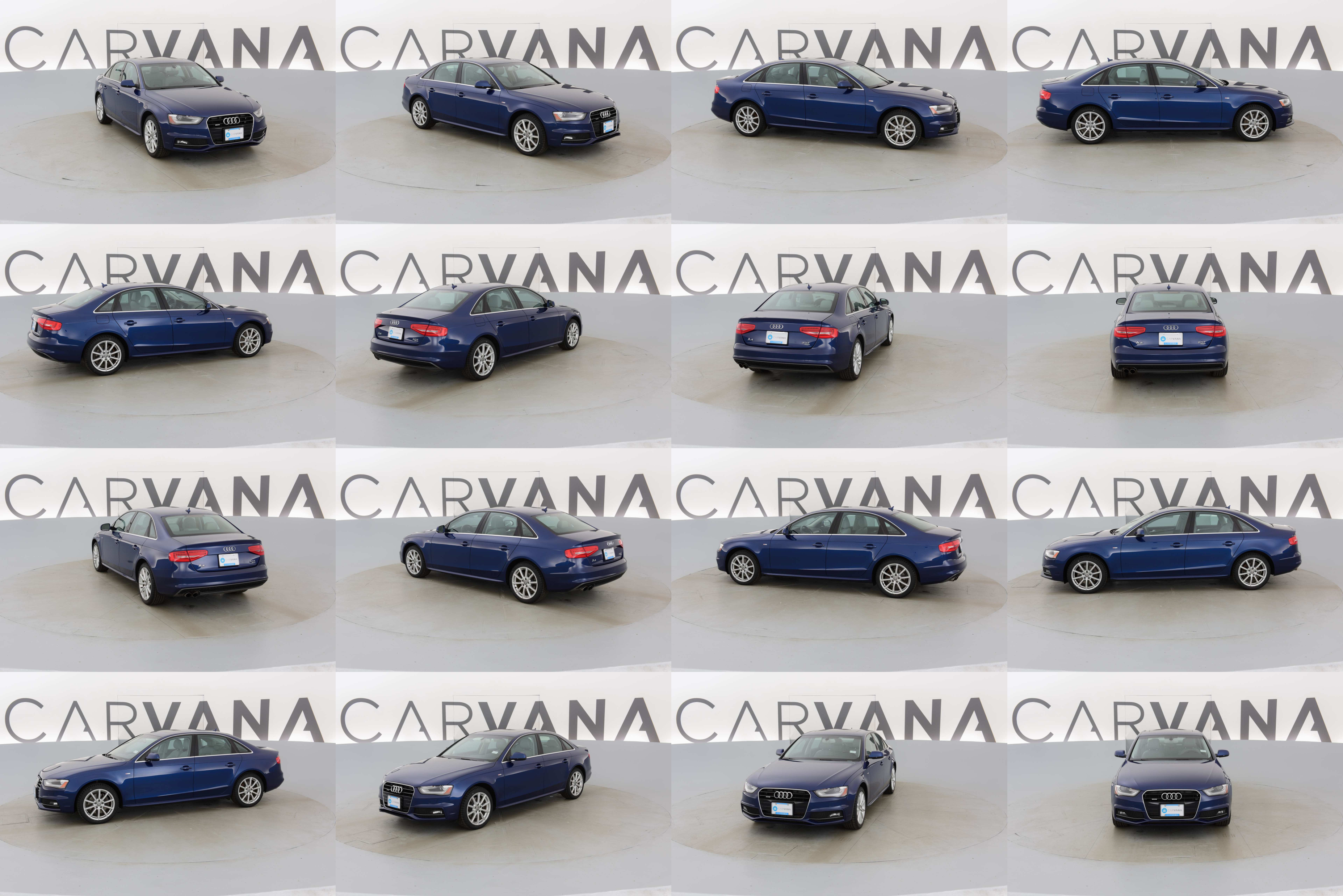}}
	\caption{Unprocessed raw data, "Carvana Image Masking Challenge" raw data example.}
	\label{data}
\end{figure}

Buying a used car is just as annoying as selling it, especially online because of the lack of complete information and transparency. Shoppers want to know everything about cars, but they must rely on often blurry pictures and little information, making online used car sales an inefficient industry. Carvana, a successful online used car startup, saw an opportunity to build long-term trust with consumers and simplify the online buying process. An interesting part of their innovation is a custom rotating photo gallery that automatically captures and processes 16 standard images of each vehicle in the inventory, a schematic diagram of which is shown in Figure \ref{data}. The dataset contains a large number of images of cars, exactly 16 images of each car taken from different angles. In the training dataset, there are 5088 images, and for each image a mask file containing a manual clipping mask is provided. While Carvana takes high-quality photos, bright reflections and cars that are similar in color to the background cause automation errors that require a skilled photo editor to make changes. Therefore, kaggle released the "Carvana Image Masking Challenge", with the aim of developing algorithms to automatically remove photo studio backgrounds.
\par 
\subsection{The proposed dataset "Local Carvana Image Masking Dataset"}
\begin{figure}[!t]
	\centerline{\includegraphics[width=\columnwidth]{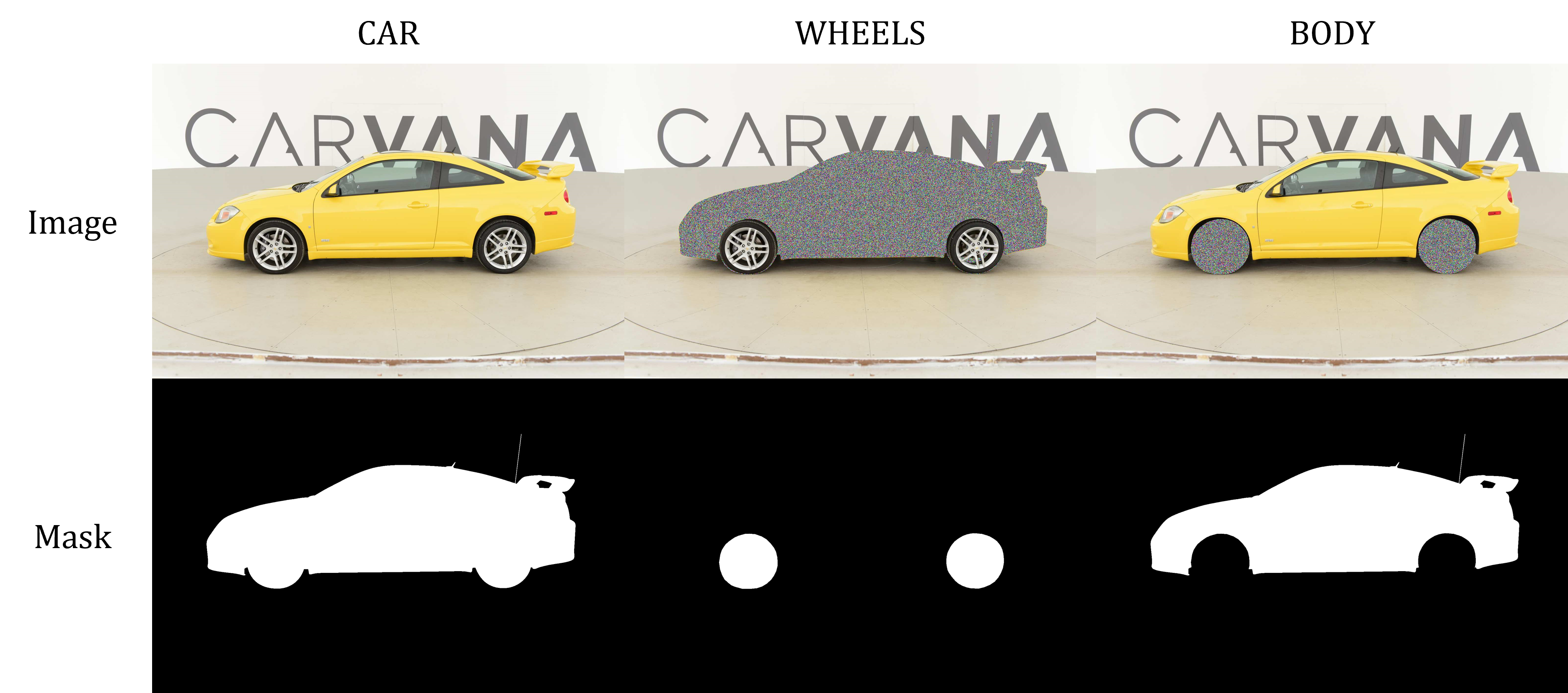}}
	\caption{Processed data, Gaussian noise is used to cover the wheels or body, and then the masked part is removed in the mask.}
	\label{fig_datas}
\end{figure}

The purpose of this work is to verify whether the local similarity of the source domain is likely to play a positive role in the transfer learning task of the target domain. It is hoped that the fewer common features between the source domains, the better, and the features extracted by the combination of source domains can cover all the features required by the target domain as much as possible. Therefore, this work selects some original images and artificially creates some locally similar source domains for use. Specifically, the original dataset used in this work is a 3-channel 8-bit RGB image in jpg format with a resolution of 1280*1918, and the value of each channel is between 0-255. We use the labeling tool PyLabelMe\cite{mpitid_2011_github} to split car images into two parts "WHEELS" and "BODY". In order to ensure that there are no common features between the source domains, we also use Gaussian noise with a mean value of 127 and a variance of 255 to cover the body and wheels of the corresponding part, and delete the corresponding part in the mask. In addition, in terms of human understanding, the actual semantic information of the source domain "WHEELS" is the wheel, and the actual semantic information of the source domain "BODY" is the body. They are completely different from each other and are only part of the target domain "CAR". The schematic diagram after processing is shown in Figure \ref{fig_datas}. We processed a total of 80 images, and split the cars in the dataset "Carvana Image Masking Challenge" into wheels "WHEELS" and body "BODY" as two source domains respectively. In addition, 40 images in the data set "Carvana Image Masking Challenge" were selected as the target domain data set of the experiment, which we named "Local Carvana Image Masking Dataset". In the experiments, no other preprocessing or data augmentation was performed on the images.
\subsection{Experimental indicators}
According to literature \cite{RN88,RN87,RN86}, the following performance metrics applicable to image segmentation are used to evaluate the performance of the proposed transfer method in the field of image segmentation, including $IOU$ score (intersection and union ratio) and ${F_\beta }$ score. Image segmentation is actually that pixels will be predicted and classified into class $c_i^{}$ after passing through the network. $TP$, $FP$, $TN$, $FN$ are four results, indicate true positive, false positive, true negative, and false negative, respectively. For the experimental index, $TP_i^{}$ represents the number of true positive samples (the number of samples correctly classified into the class $c_i^{}$). $FP_i^{}$ represents the number of false positive samples (the number of samples that are misclassified into class $c_i^{}$). $TN_i^{}$ represents the number of true negative samples (the number of samples correctly classified as non-class $c_i^{}$ samples). $FN_i^{}$ represents the number of false negative samples (the number of samples that are misclassified as non-class $c_i^{}$ samples).
\par 
$IOU$ can be understood as the ratio of the intersection and union of the real correct area and the predicted correct area. The $IOU$ score $IOU_i$ of class $c_i^{}$ is calculated as follow:
\begin{equation}
IOU_i^{} = \frac{{TP_i^{}}}{{FP_i^{} + FN_i^{}}}\label{eq12}
\end{equation}
In addition to the $IOU$ score indicator, $Precision$ score and $Recall$ score are two commonly used basic indicators. The $Precision$ score is the ratio of the number of correctly predicted true positive pixels to the total number of predicted positive pixels (from the perspective of prediction results, how many predictions are accurate). The $Recall$ score is the proportion of predicted correct true positive pixels to the total number of true positive pixels (from the perspective of true labeling, how many are recalled). The $Precision$ score $Precision_i$ and $Recall$ score $Recall_i$ of the class $c_i^{}$ are defined as follows:
\begin{equation}
Precision_i^{} = \frac{{TP_i^{}}}{{TP_i^{} + FP_i^{}}}\label{eq13}
\end{equation}
\begin{equation}
Recall_i^{} = \frac{{TP_i^{}}}{{TP_i^{} + FN_i^{}}}\label{eq14}
\end{equation}
Sometimes the target task needs to consider the $Precision_i$ score and the $Recall_i$ score comprehensively. At this time, the ${F_\beta }$ score comes into being. Its calculation method is as follows:
\begin{equation}
\begin{array}{l}
	F_{i^{}_\beta } = \left( {1 + {\beta ^2}} \right) \times \frac{{Precision_i^{} \times Recall_i^{}}}{{\left( {{\beta ^2} \times Precision_i^{}} \right) + Recall_i^{}}}\\
	= \frac{{\left( {1 + {\beta ^2}} \right) \times TP_i^{}}}{{\left( {1 + {\beta ^2}} \right) \times TP_i^{} + {\beta ^2} \times FN_i^{} + FP_i^{}}}
\end{array}\label{eq15}
\end{equation}
$\beta$ can take different values according to the focus of the task index. When $\beta$ takes $1$, it is $F_1$ score, that is the famous $Dice\;score$.
The above test indicators are only for specific images in the data set. To reflect the situation of the entire data set, it is necessary to test all the pictures, and then take the mean and standard deviation.
\subsection{Experiment description}
The source domain used in this paper is two completely different objects, which are only components of the target domain. For this reason, the pixels of the segmented area are regarded as positive samples to ensure that the local features are complementary. All source domain images are used as the training set. Since the usual assumption in transfer learning is that there is less data in the target domain, and we hope that as much data as possible is used in the verification and testing phases to obtain accurate results. Therefore, the target domain images are randomly divided into training set, validation set and test set according to the ratio of $6:2:2$, as shown in Table 1. The segmentation task model is selected as U-Net, and the remaining parameters are listed in Table \ref{sample}.
\begin{table}[]
	\centering % 居中
	\caption{The sample size in the datasets.}   
	\label{sample}
	\begin{tabular}{llll}
		\hline
		Data Set        & Training set & Validation set & Test set \\ \hline
		CAR             & 24           & 8              & 8       \\
		WHEELS          & 80           & -              & -        \\
		BODY            & 80           & -              & -        \\ \hline
	\end{tabular}
\end{table}
\par 
MSATL represents the proposed Multi-source Adversarial Transfer Learning method. MSATL* is an ablation experiment, which means removing the attention mechanism used to select local features. MSATL** is an ablation experiment, denoting deletion of the multi-source domain independent strategy. DG means that Domain Generalization\cite{RN82} is a method of multi-source transfer learning using domain generalization mode\cite{RN26,RN49}. The most common features in different source domains will be extracted through confrontational thinking. MMI is the Multi-model integration method of work\cite{RN42}. The difference from this work is that this method trains the model separately for each source, and then implements transfer learning by selecting a classifier with higher accuracy. MBN represents the multi-branch network feature fusion mode of the work\cite{RN50} structure for multi-source transfer learning. The segmentation task model is selected as U-Net, and the remaining parameters are listed in Table \ref{Parameters_setting}, and all methods share the same experimental data and settings.
\begin{table}[]
	\centering % 居中
	\caption{Parameters setting}   
	\label{Parameters_setting}
\begin{tabular}{ll}
	\hline
	Parameter                  & Parameter     \\ \hline
	$\lambda$				   & 0.3           \\
	Random seed                & 0             \\
	Epochs                     & 100           \\
	Sub-batch size             & 2             \\
	Learning rate              & $1 \times 10_{}^{ - 3}$              \\
	Optimizer                  & Adam\cite{RN81}  \\
	Segmentation loss function & Cross-Entropy \\ \hline
\end{tabular}
\end{table}
\subsection{Results and Analysis}
For the car segmentation task, since the two source domains, wheels and body, are different types of objects, our main work is to lower the multi-source transfer learning to the local feature level, and at the same time filter the local features to avoid forced matching of inappropriate features. To this end, for the exploration of interpretability in this work, the features extracted from each sub-network are visualized in the form of a heat map, where the closer the color is to red, the greater the weight of the features extracted from this part. The closer to blue, the smaller the weight of the feature extracted from this part, as shown in Figure 8, it can be seen that when BODY is used as the source domain, the weight of the features extracted by the sub-network from the body is larger. When WHEELS is used as the source domain, the weight of the wheel in the features extracted by the sub-network is relatively large. This is consistent with the observation results of the human eye.
\begin{figure}[!t]
	\centerline{\includegraphics[width=\columnwidth]{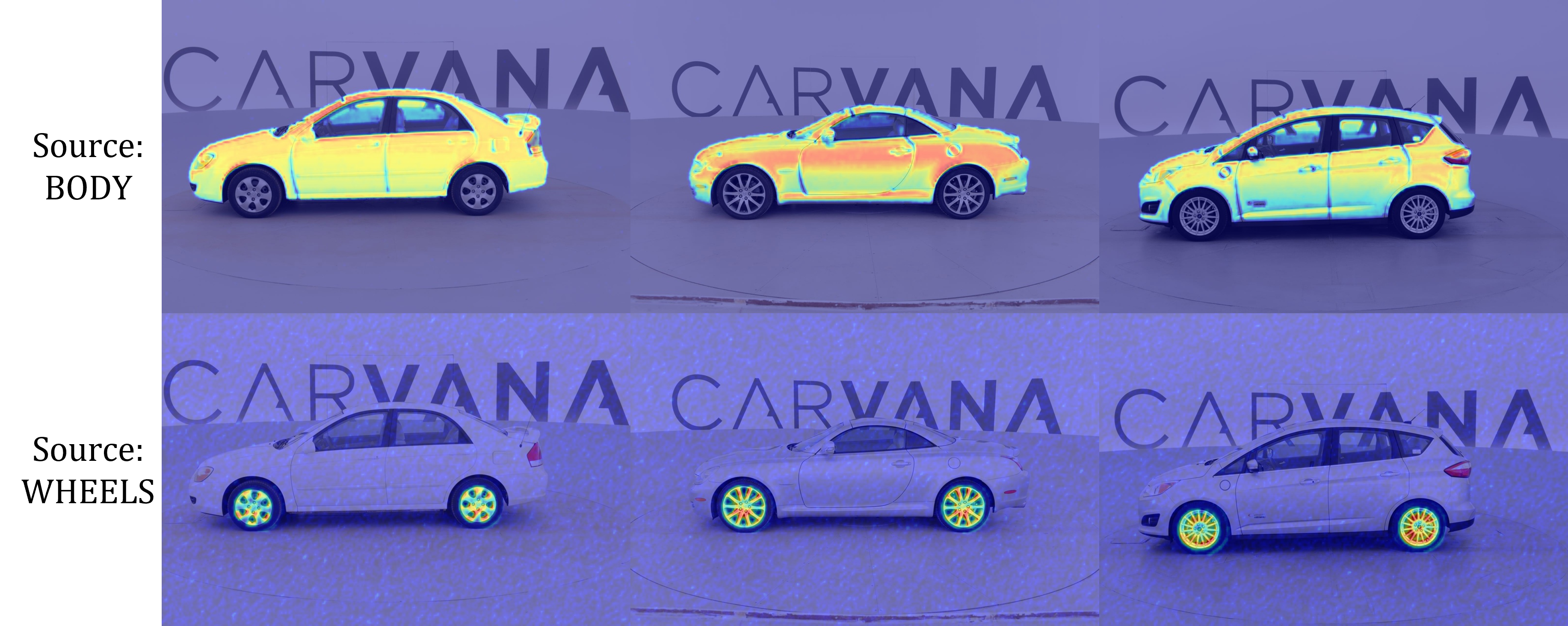}}
	\caption{Visualization of sub-network features corresponding to different source domains.}
	\label{v}
\end{figure}
We also compare the proposed method with some other thought methods and visualize the segmentation results as Figure \ref{results}. Specifically, as shown in Table \ref{result}, the proposed MSATL segmentation task achieves the best transfer performance, its $IOU$ and $Dice score$ are both the highest, moreover, their standard deviation is also the smallest;
The ablation experiment MSATL* proves the effectiveness of the attention mechanism for local feature selection and segmentation tasks. Its $IOU$ and $Dice score$ indicators are $0.90910$ and $0.95128$, respectively, and their standard deviations are greater than MSATL;
The ablation experiment MSATL** proves the effectiveness of the multi-source domain independent strategy for segmentation tasks. Its $IOU$ and $Dice score$ are $0.90876$ and $0.95197$, respectively. It is worth noting that their standard deviation $(std)$ is more than $10$ times that of MSATL.
Current common multi-source transfer learning strategies are not as good as the proposed method and its ablation experiments. Among them, the performance of DG is the weakest, because the idea of DG is to find the common features of all domains, while the common features of wheels and body are limited, resulting in insufficient features. MMI performance is relatively strong, because the method is based on the idea of source domain selection. For this task, the body BODY has more pixels. Even if the model of this source domain is selected, it is possible to obtain relatively acceptable results. MBN is worse than DG, but it is worth noting that its $std$ value is larger, indicating that the model is not as stable as DG and MMI. We believe that the assumption of this method is that the underlying features are all transferable, and their high-level feature combinations also have the possibility of fusion, while in fact the common features of car body and tire are limited, such as their contours and textures are quite different.
\begin{figure}[!t]
	\centerline{\includegraphics[width=\columnwidth]{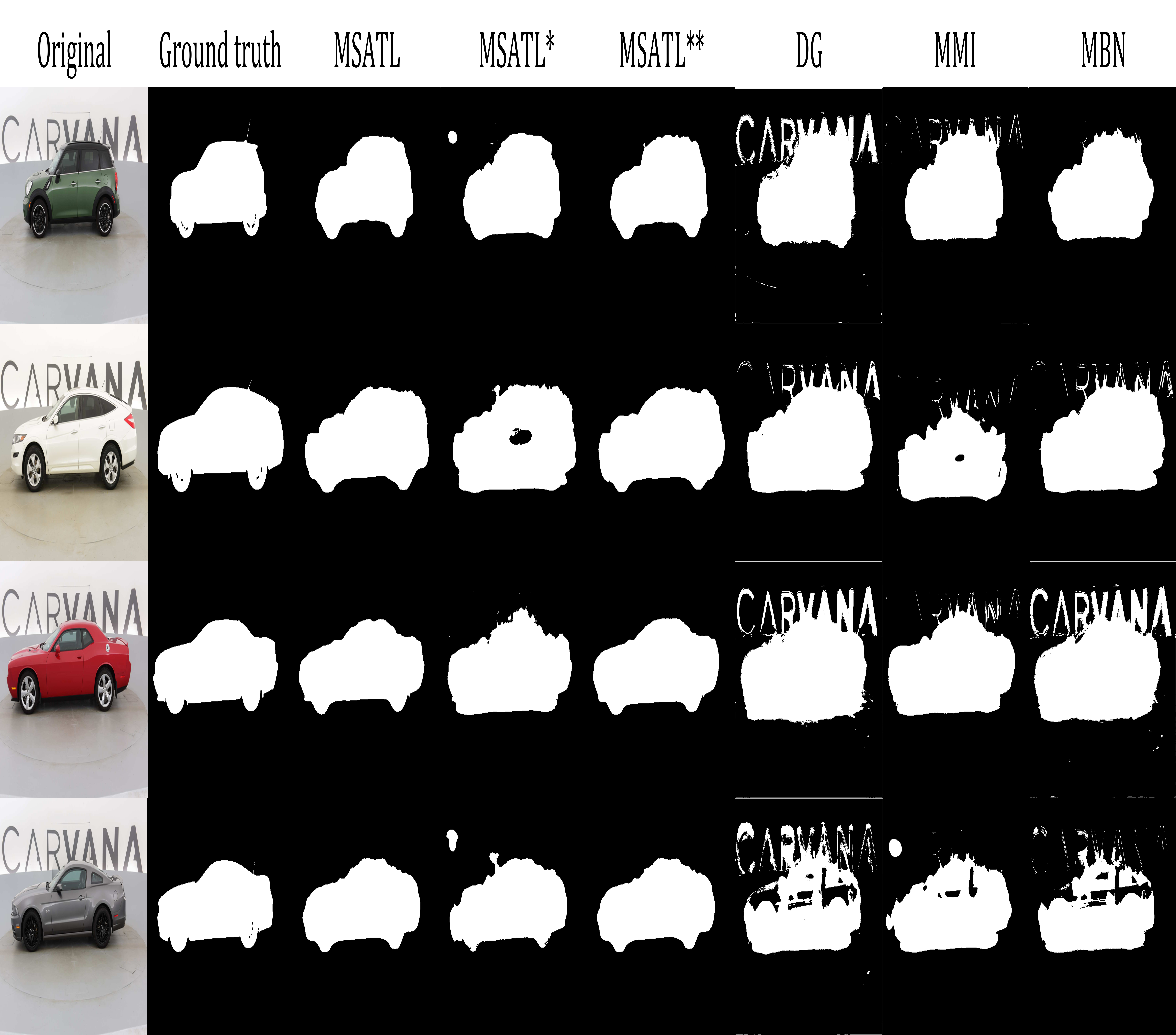}}
	\caption{Visual comparison of segmentation experiment results of different transfer learning methods on the proposed dataset "Local Carvana Image Masking Dataset".}
	\label{results}
\end{figure}
\begin{table}[]
	\centering % 居中
	\caption{Segmentation experimental results of different transfer learning methods on the proposed dataset "Local Carvana Image Masking Dataset"}   
	\label{result}
	\begin{tabular}{lllll}
		\hline
		Methods & IOU     & IOU std & Dice Score & Dice Score std \\ \hline
		MSATL   & 0.94827 & 0.00395 & 0.97344    & 0.00208        \\
		MSATL*  & 0.9091  & 0.00967 & 0.95128    & 0.00548        \\
		MSATL** & 0.90876 & 0.03985 & 0.95197    & 0.02303        \\
		DG      & 0.80437 & 0.01968 & 0.84184    & 0.01324        \\
		MMI     & 0.83242 & 0.018   & 0.85539    & 0.0122         \\
		MBN     & 0.80562 & 0.03814 & 0.84212    & 0.02364        \\ \hline
	\end{tabular}
\end{table}

\section{CONCLUSION}\label{CONCLUSION}
Aiming at the scenario where only some local feature similarities can be provided, but the global feature similar source domain data cannot be provided, a Multi-source Adversarial Transfer Learning method based on local feature similar source domains is proposed. The method extracts transferable features between locally similar domains through multiple local feature extraction networks, and utilizes an attention mechanism to strengthen transferable local features and suppress mismatched features. This method designs a multi-source domain independent strategy, and makes the target domain data of each batch exactly the same through constraints, so as to fully extract local features that can be fused.
\par 
The transferability of the method is verified on the proposed dataset "Local Carvana Image Masking Dataset". Experimental results show that the proposed method has better performance in the case of limited overall similarity but local feature similarity. Compared with the existing multi-source transfer learning ideas, the proposed method can be well applied in this special scenario. Overall, this method can effectively extract and mine knowledge in multiple source domains with limited overall similarity but certain similarity in local features, and then fuse them. It has wide applicability when the types of data sets are limited.
\par 
Our method uses the idea of adversarial, which provides the possibility to apply to the scene where the target domain has a large amount of data but lacks labels. In the future, we will explore the feasibility of this method in this scenario. In addition, we also consider the relationship between multiple attention modules, such as designing rules between multiple attention modules to make the extracted features more complementary. Most importantly, we also hope to further explore the interpretability of multi-source transfer learning with similar local features. Unlike the current selection of source domains based on the appearance of the naked eye, we hope to design a quantitative local similarity source domain. The selection method makes the selection of the source domain more reasonable and well-founded.

\section*{Acknowledgements}

This work was supported in part by the National Natural Science Foundation of China (61973067).

%\section*{CRediT author statement}

%Yifu Zhang: Conceptualization, Methodology, Software, Validation, Formal analysis, Writing - Original Draft.\\Hongru Li: Supervision, Project administration, Funding acquisition, Writing - Review \& Editing. \\Tao Yang: Writing - Review \& Editing. \\Jan Jansen: Supervision.: Ajay Kumar: Software, Validation.: Sun Qi: Writing- Reviewing and Editing,

\bibliography{wpref}

\end{document}